\documentclass[letterpaper, 10 pt, conference]{ieeeconf}  

\usepackage{algorithm, algpseudocode, graphicx,float, dsfont, amsmath,amssymb,cancel,arydshln,bm, etoolbox}
\usepackage{graphicx,amsmath,amssymb,bm,xcolor,soul}
\usepackage{color, colortbl}
\usepackage{multirow}
\usepackage{multicol}
\usepackage{blindtext}
\usepackage{hyperref}
\usepackage{todonotes}
\usepackage[font=small,labelfont=bf]{caption}
\graphicspath{ {./figures/} }

\patchcmd\subequations
 {\theparentequation\alph{equation}}
 {\subequationsformat}
 {}{}

\newcommand{\subequationsformat}{\theparentequation.\alph{equation}}
\newcommand{\trsp}{{\scriptscriptstyle\top}}
\newcommand{\RNum}[1]{\uppercase\expandafter{\romannumeral #1\relax}}

\IEEEoverridecommandlockouts                              

\title{\LARGE \bf
Learning Joint Space Reference Manifold\\for Reliable Physical Assistance
}

\author{Amirreza Razmjoo$^{1,2}$, Tilen Brecelj$^{3}$, Kristina Savevska$^{3}$, Ale\v{s} Ude$^{3}$, Tadej Petrič$^{3}$, and Sylvain Calinon$^{1,2}$
\thanks{*This work was supported by the SWITCH project (\url{https://switch-project.github.io/}), funded by the Swiss National Science Foundation and Slovenian Research Agency with grant number N2-0153.}
\thanks{\textsuperscript{1} Idiap Research Institute, CH {\tt\small\{{name.surname\}}@idiap.ch}}%
\thanks{\textsuperscript{2} \'Ecole Polytechnique F\'ed\'erale de Lausanne (EPFL), CH}%
\thanks{\textsuperscript{3} Jožef Stefan Institute, SI {\tt\small\{{name.surname\}}@ijs.si}}%
}

\usepackage{etoolbox}
\makeatletter
\patchcmd{\@makecaption}
  {\scshape}
  {}
  {}
  {}
\makeatother

\begin{document}
\maketitle
\thispagestyle{empty}
\pagestyle{empty}
\begin{abstract}
This paper presents a study on the use of the Talos humanoid robot for performing assistive sit-to-stand or stand-to-sit tasks. In such tasks, the human exerts a large amount of force (100--200 N) within a very short time (2--8 s), posing significant challenges in terms of human unpredictability and robot stability control. To address these challenges, we propose an approach for finding a spatial reference for the robot, which allows the robot to move according to the force exerted by the human and control its stability during the task. Specifically, we focus on the problem of finding a 1D manifold for the robot, while assuming a simple controller to guide its movement on this manifold. To achieve this, we use a functional representation to parameterize the manifold and solve an optimization problem that takes into account the robot's stability and the unpredictability of human behavior. We demonstrate the effectiveness of our approach through simulations and experiments with the Talos robot, showing robustness and adaptability.
\end{abstract}

\section{INTRODUCTION}
\label{sec:intro}
The integration of robots into our daily lives is becoming increasingly feasible, and their presence in society is no longer a distant possibility. This possesses the capability to provide different forms of assistance to individuals in their daily lives, including social \cite{rasouli_potential_2022}, collaborative \cite{Scassellati_2022}, and physical support \cite{Ajoudani_2022}. The current study specifically focuses on physical assistance, in which a humanoid robot, Talos, aids users in performing sit-to-stand or stand-to-sit (STS) tasks. 

Previous research in STS tasks has mainly concentrated on utilizing optimization techniques to estimate optimal assistance, with the aim of solving the problem offline \cite{Geravand2017,Razmjoo21ICAR,Li2021}. The experimental setup utilized in these studies involved low-dimensional, fully actuated systems, allowing for the admissible implementation of pre-planned assistance in the reproduction phase. In this scenario, the human acts as a follower system, adapting his/her behavior to the pre-planned actions of the robot. However, this paper aims to achieve a more human-like STS task, where the robot plays the role of a follower and reacts to the human's actions while considering different constraints. A similar study to our work is the one conducted by Shimon et al. \cite{shimon_STS_2015}, where a ballbot was used to assist a human in an STS task by controlling its leaning angle based on the force received from the human.  

\begin{table}[t]
\begin{center}
    \begin{tabular}{cc}
        \centering
        \includegraphics[width =0.35\columnwidth]{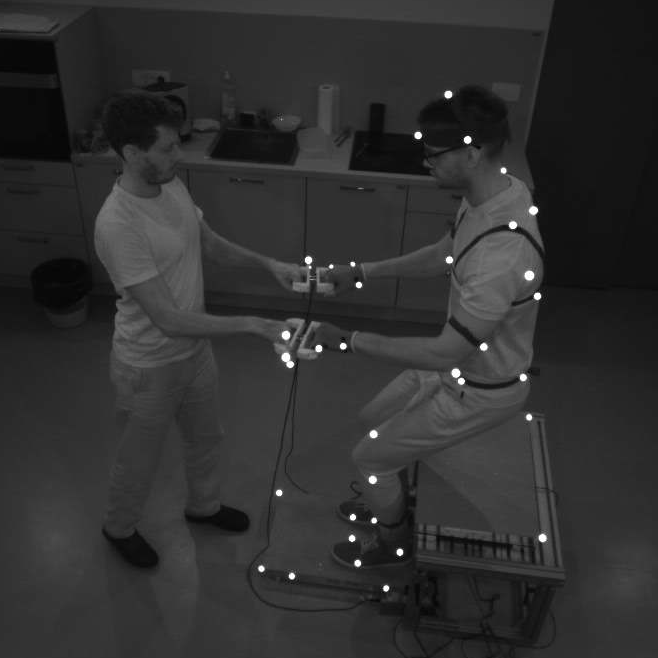}&
        \includegraphics[width=0.45 \columnwidth]{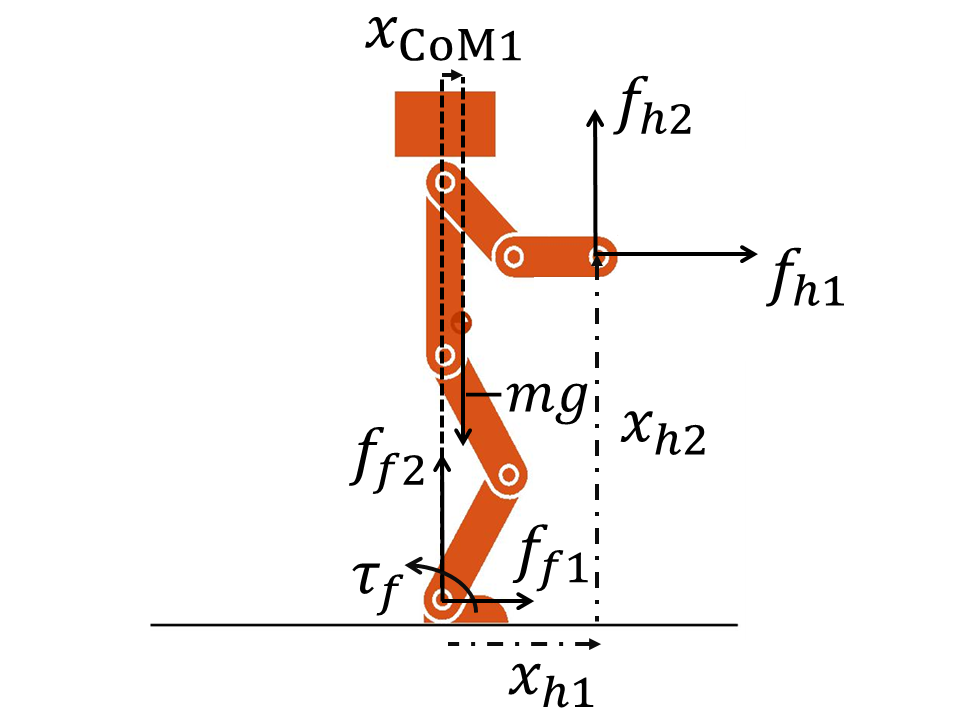} \\
        a) Human-human interaction & b) Dynamic model\\
        \includegraphics[width =0.35\columnwidth]{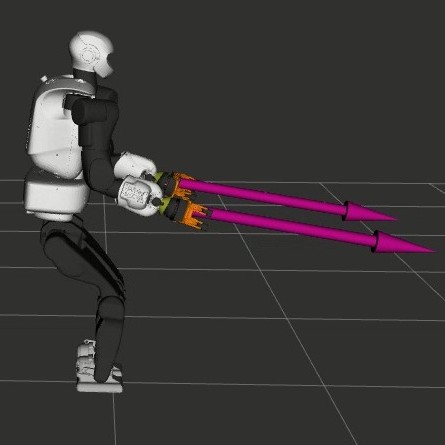} &
       \includegraphics[width =0.35\columnwidth]{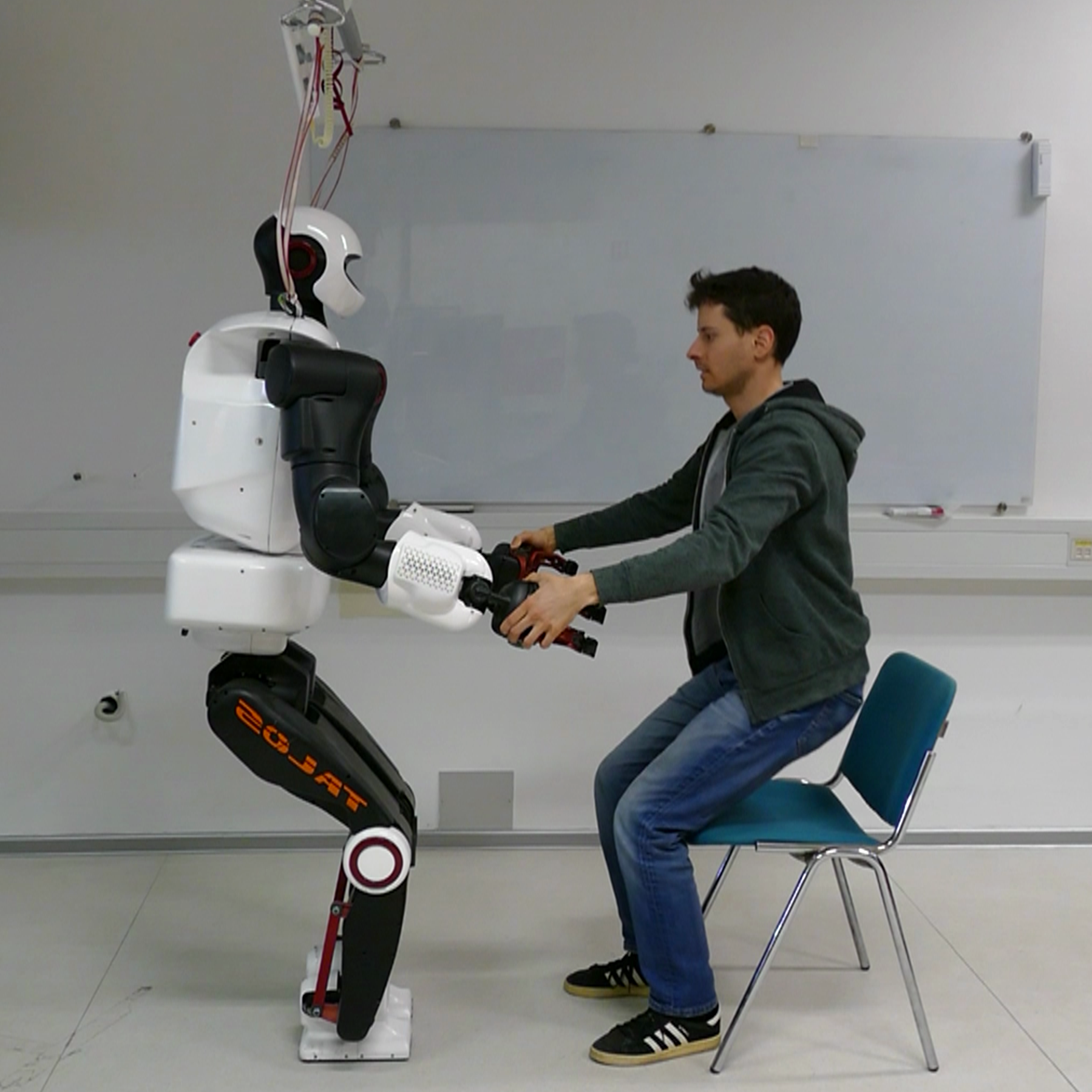} \\ 
       c) Simulation & d) Real robot
    \end{tabular}     
    \captionof{figure}{Experimental setup. (a) Force and human articulation position data collection during human-human interaction (only force data is utilized in this paper), (b) Robot's dynamic model, (c) Simulation setup where the human is modeled as an external force, and (d) human-robot interaction scenario.}\vspace{-5mm}
    \label{fig:experimental_setup}
\end{center}
\vspace{-0.3cm}
\end{table} 

Humanoid robots hold a distinct position among various types of robots, as their physical resemblance to humans has fueled aspirations that humanoid robots may one day perform tasks at a level comparable to humans. However, these robots currently face numerous challenges, including underactuated dynamics, high dimensionality, and unilateral contact with the environment. Overcoming these research challenges has led to the development of reduced-order models such as the linear inverted pendulum mode (LIPM) \cite{Kajita1991} or centroidal dynamics \cite{dai2014whole}, to reduce the dimensionality of the system. Additionally, different strategies such as hip or ankle strategies have been developed to recover the robot from unstable states \cite{Atkeson2007}. Optimization-based methods have been employed to consider the locomotion of the robot while planning for the unactuated states and constraints \cite{dai2014whole,Ibanez2018}. Another approach is to partition the robot into multiple kinematic chains, with some of them focusing on task performance and others being responsible for balance maintenance \cite{Ibanez2012}.

In the context of physical interactions between humanoid robots and humans, such as STS tasks, the aforementioned issues to control a humanoid become significantly more critical. STS tasks are challenging as they require humans to exert substantial force in a short period. According to data published in \cite{shimon_STS_2015} and as demonstrated in this paper, the applied force can range between 100--200 N within 2--8 seconds. Consequently, the robot must respond promptly to these forces while maintaining its stability and adhering to all constraints. Optimization-based approaches utilized in other studies are mostly impractical for this problem as they can be time-consuming and they usually rely on accurate environmental models that are unfeasible here due to the human unpredictability.

One approach to limit these issues can be to restrict the robot's motion to a manifold, enabling the robot to avoid solving a high-dimensional problem, as its motion is now constrained to exist solely on the manifold. It also enables the robot to exclude certain constraints, such as joint limits, from its consideration, given that these constraints have already been accounted for in the manifold.

Reduced models, such as centroidal dynamics, can also offer a promising solution. However, a critical question remains challenging: how to convert center-of-mass (CoM) motion into joint configurations while maintaining system stability. Considering the robot and task redundancy, one potential solution is to utilize methods such as the centroidal momentum matrix presented in \cite{orin_centroidal_2013} to map the robot's CoM acceleration back to the joint acceleration by solving a regression problem. Constraints or other tasks can be considered in the null space of this problem. Unfortunately, this method cannot guarantee that all constraints will be met. Furthermore, the regression approach used in these methods can lead to unintended behaviors if the user behaves unexpectedly. Restricting the inverse kinematic problem to a manifold can alleviate certain challenges. Specifically, kinematic constraints become redundant, as the relevant information is already incorporated within the manifold. Additionally, this reference can be computed and pre-evaluated offline, thereby it ensures that the robot's trajectory remains constrained within the manifold, regardless of the user's interactions with the robot.

Motivated by the objective of employing a manifold, this paper tackles the problem of robot control by addressing a fundamental question of how to find this manifold for physical STS assistance. Some simplifying assumptions are made to enable the robot to control its stability by regulating its CoM exclusively in the anterior/posterior direction, similar to the approach taken in \cite{shimon_STS_2015}. This leads to a one-degree-of-freedom problem which motivates us to find a robust correlation among the redundant robot joint configurations as a manifold to enable the robot to exhibit robust behavior in the face of unpredictable human actions. The paper demonstrates the effectiveness of this approach through simulation experiments involving various human behaviors, where a real-world robotic experiment is conducted to verify the feasibility of the proposed method.

\section{BACKGROUND}
\label{sec:background}
\subsection{Robot Stability}
Different approaches have been proposed to measure the robot stability, such as the Zero Moment Point (ZMP) \cite{Kajita1991}, which is also known as the Center of Pressure (CoP), and the Contact Wrench Cone (CWC) \cite{Hirukawa_2006}. In this work, we focus on the ZMP, which is the point on the ground where the horizontal moment applied on the robot is zero. For the 2D robot depicted in Fig. \ref{fig:experimental_setup}, this point can be calculated as
\begin{equation}\label{eq:zmp}
    x_{\text{ZMP}} = \frac{{\tau}_{\text{f}}}{f_{\text{f2}}} \!\underset{}{\overset{\text{static cond.}}{\rightarrow}}\! x_{\text{ZMP}} = \frac{f_{\text{h1}}x_{\text{h2}}\! -\! f_{\text{h2}}x_{\text{h1}} \!-\! mgx_{\text{CoM1}}}{-mg - f_{\text{h2}}},
\end{equation}
where $g = -9.81\, \frac{\text{m}}{\text{s}^2}$ is the gravity acceleration. According to the data we have received from the human side (Fig.~\ref{fig:force_hh}), $f_{\text{h2}}x_{\text{h1}} \ll f_{\text{h1}}x_{\text{h2}}$ and $f_{\text{h2}} \ll mg$, so \eqref{eq:zmp} can be simplified to
\begin{equation}\label{eq:zmp_sim}
    x_{\text{ZMP}} \approx \frac{f_{\text{h1}}x_{\text{h2}} - mgx_{\text{CoM1}}}{-mg}.
\end{equation} 

For the robot to be stable, $x_{\text{zmp}}$ should be located inside the support polygon. For the 2D case, this support polygon can be denoted as $\delta^{-} \leq x_{\text{ZMP}} \leq \delta^{+},$ where $\delta^{\pm}$ correspond to the two ends of the foot location on the ground. 

\subsection{Functional Representation}\label{sec:MP}
Parameterizing a trajectory as a functional representation, such as motion primitives, has become a widely used approach in the field of Learning from Demonstration (LfD), as evidenced by the popularity of methods such as Dynamical Movement Primitives (DMP) \cite{Ijspeert2013} or Probabilistic Movement Primitives (ProMP) \cite{paraschos_probabilistic_2013}, see \cite{Calinon19MM} for an overview of using basis functions for motion parameterization. This technique can reduce the number of parameters to describe a movement while ensuring its smoothness, which is often difficult to achieve otherwise. Due to these advantages, they have also proven to be useful in optimal control problem formulation \cite{Razmjoo21ICAR}.

Consider a $d$-dimensional variable $\bm{q}(s)$ on a 1D manifold, where $s \in [0,1]$ is a mapping variable that indicates the relative position of the point on the manifold. It is assumed that each element $q_i(s)$ of $\bm{q}(s)$ where $i \in \{1,2,\hdots, d\}$ can be expressed as a weighted combination of $n$ basis functions $q_i(s) = \bm{\phi}(s) \bm{w}_i$, where $\bm{\phi}(s) \in \mathbb{R}^{1 \times n}$ is the vector of basis functions at the location $s$, and $\bm{w}_i$ is the corresponding weighting vector.

By concatenating all dimensions, we can obtain a similar formulation $\bm{q}(s) = \bm{\Psi}(s) \bm{w}$ with $\bm{\Psi}(s)=\bm{\phi}(s) \otimes \bm{C}$, where $\otimes$ is the Kronecker product, $\bm{C} \in \mathbb{R}^{d \times r}$ is a coordination matrix, and $\bm{w}$ is the concatenated weighting vector. Often, $\bm{C} = \bm{I}_d$ is used as coordination matrix to decouple the $d$ dimensions. However, alternative selections can be made to establish correlations among the variables. 

Another approach to describe a manifold is to represent this with $K$ discrete points uniformly spread on the manifold. Compared to this method, the functional representation approach not only reduces the dimensionality of the path from $d \times K$ to $r \times n$ (where $n \ll K$ and $r \leq d$) but also ensures smoothness due to the continuity of the basis functions. In this paper, Bernstein polynomials \cite{FAROUKI2012379} were chosen as basis functions (also used for Bézier curves), although other alternatives such as Radial Basis Functions (RBFs) can also be used.

\section{Methodology}
\label{sec:method}
In this section, we introduce a set of assumptions to simplify the problem, followed by our objective to construct a manifold. The manifold can be regarded as a lookup table whose smoothness implementation on the actual robot is ensured by exploiting the functional representation idea. 
\subsection{Assumptions}
To simplify the problem, several assumptions are made:
\begin{enumerate}
    \item The robot will not move its feet during the task, meaning that no stepping action is required.
    \item  The two feet of the robot are located in the same position when viewed from the sagittal plane and the forces applied to the robot's hands are identical, resulting in symmetrical behavior on the left and right sides of the robot. This correlation can be formulated through the coordination matrix $\bm{C}$ presented in Sec.~\ref{sec:MP}.
    \item We control only the ankle, knee, hip, shoulder (3 DoF), and elbow joints, which are referred to as active joints $\bm{q}^{\text{active}} \in \mathcal{R}^{14}$ hereafter.
    \item The robot can apply sufficient effort to achieve a desired behavior. 
\end{enumerate}

The first two assumptions allow the problem to be simplified by controlling only the robot CoM in the anterior/posterior direction, similarly as in \cite{shimon_STS_2015}. The fourth assumption allows the problem to not consider the actuator limits during the planning phase which is a common assumption when working with humanoids \cite{wieber2016modeling, dai2014whole}.

\subsection{Deriving a Spatial Reference}
The problem addressed in this paper can be formulated as finding a suitable configuration for the humanoid robot to maintain its balance while being subjected to external forces from a human in a collaborative task, namely
\begin{equation}\label{eq:main_problem}
\begin{gathered}
    \ddot{\bm{Q}}^* = [\ddot{\bm{q}}^*_{0}, \ddot{\bm{q}}^*_{1}, \hdots, \ddot{\bm{q}}^*_{T}] = \underset{\ddot{\bm{Q}}}{\arg\min} \sum_{t=0}^{t = T} c_t(\bm{q}_t, \bm{f}_t, \bm{u}_t), \\
    \text{s.t.}  \ \bm{q}(0) = \bm{q}_0,  \ \bm{q}_t \in \mathcal{Q}^{\text{active}}, \ \bm{u}_t \in \mathcal{U}^{\text{active}}, \\ 
     \ \text{DY}(\bm{q}_t, \dot{\bm{q}}_t, \ddot{\bm{q}}, \bm{f}_t, \bm{u}_t)=0, \ g(\bm{q}_t, \dot{\bm{q}}_t, \ddot{\bm{q}}, \bm{f}_t, \bm{u}_t) \leq 0 \\ \text{ZMP}(\bm{q}_t, \dot{\bm{q}}_t, \ddot{\bm{q}}, \bm{f}_t, \bm{u}_t) \in \delta^{\pm} , \ \ \ \forall t \in \{0,\hdots, T\}, 
\end{gathered}
\end{equation}
where for each time step $t$, the cost function is defined as $c_t$, the force applied at the robot's hands is denoted as $\bm{f}_t$, and the control command as $\bm{u}_t$. The dynamic equation of the system is represented by the function $\text{DY}(\cdot)$, while the ZMP point of the system is calculated using the function $\text{ZMP}(\cdot)$. Other constraints that may be necessary for the task are incorporated through the function $g(\cdot)$. The feasible set for active joint angles and their control commands are denoted by $\mathcal{Q}^{\text{active}}$ and $\mathcal{U}^{\text{active}}$, respectively.

To incorporate feedback from the environment or to adapt to different humans, the problem represented by \eqref{eq:main_problem} must be solved recurrently using a method such as model predictive controller (MPC). However, this approach does not seem suitable for pHRI tasks as the environmental model can vary significantly at each time step. Instead of solving \eqref{eq:main_problem} for the whole trajectory, our focus is to find a desired terminal configuration $\bm{q}_T$ at each time, which is expected to be static at the end of the task, i.e., $\dot{\bm{q}}_T = \ddot{\bm{q}}_T = \bm{0}$. Therefore, we can assume static motion and use the formula derived for the ZMP \eqref{eq:zmp_sim}, which allows us to define the ZMP as a function of the current configuration $\bm{q}$ and the horizontal external force applied by the user $f_{\text{h1}}$, namely
\begin{equation}\label{eq:target_problem}
\begin{gathered}
    \bm{q}^*_T = \underset{\bm{q}_T}{\arg\min} \ \ c(\bm{q}_T, f_{\text{h1}}), \\
    \text{s.t.}  \ \ \bm{q}_T \in \mathcal{Q}^{\text{active}}, \ \text{ZMP}(\bm{q}_T,f_{\text{h1}}) \in \delta^{\pm}, \ g(\bm{q}_T,f_{\text{h1}}, f_{\text{h2}}) \leq 0.
\end{gathered}
\end{equation}
For the sake of enhancing readability, the notation of $T$ is omitted hereafter. The cost function $c$ should be defined to consider the goal of the task which is the stability of the robot. One potential approach for realizing this objective involves defining the cost function as 
\begin{equation}\label{eq:standard_cost}
    c_{\text{std}}(\bm{q},f_{\text{h1}}) = \|\text{ZMP}(\bm{q},f_{\text{h1}})\|^2.
\end{equation}

However, due to the unpredictable nature of human behavior, the desired configuration should be suitable for a wide range of possible force inputs. Considering this unpredictability seems to be a very crucial point that is usually overlooked in previous research \cite{Geravand2017,shimon_STS_2015} due to their low-dimensional or fully actuated robotic systems. Therefore, an alternative approach to defining the cost function is
\begin{multline}\label{eq:cost_function}
c_{\text{rob}}(\bm{q}) = \sum_{p = 0}^{p = f_{\text{max}}} {\Big\|\text{ReLU}\big(\text{ZMP}(\bm{q},p) - \delta^{+}\big)\Big\|}^2 \\ 
+ {\Big\|\text{ReLU}\big(\delta^{-} - \text{ZMP}(\bm{q},p)\big)\Big\|}^2,
\end{multline}
where $f_{\text{max}}$ is the maximum force that we expect to receive from the user. This cost penalizes the system quadratically according to how much it has violated the boundaries. 

\subsection{Functional Representations}
Combining \eqref{eq:cost_function} and \eqref{eq:main_problem} will help us obtain a robust configuration of the robot that specifies where the robot should be according to different forces (e.g., using a lookup table). However, solving this problem in its current form can have some drawbacks. For example, we need to solve \eqref{eq:main_problem} for each value of $f_{\text{h1}}$, which can result in a jump between different solutions, see Fig.~\ref{fig:different_steps}.a. The aforementioned behavior can potentially influence the quality of human-robot interaction, in addition to affecting the sensory information, particularly the force values, if the robot experiences shaking during its operation. To overcome this limitation, we deploy the idea of a functional representation as it can correlate the solution for different force values. Specifically, we divide the problem into two parts by introducing a new intermediate variable $s \in [0,1]$ and assuming that the desired configuration is composed of some basis functions that are functions of $s$ such that

\begin{table}[t]
\begin{center}
    \begin{tabular}{cc}
        \centering
        \includegraphics[width = 0.45\columnwidth]{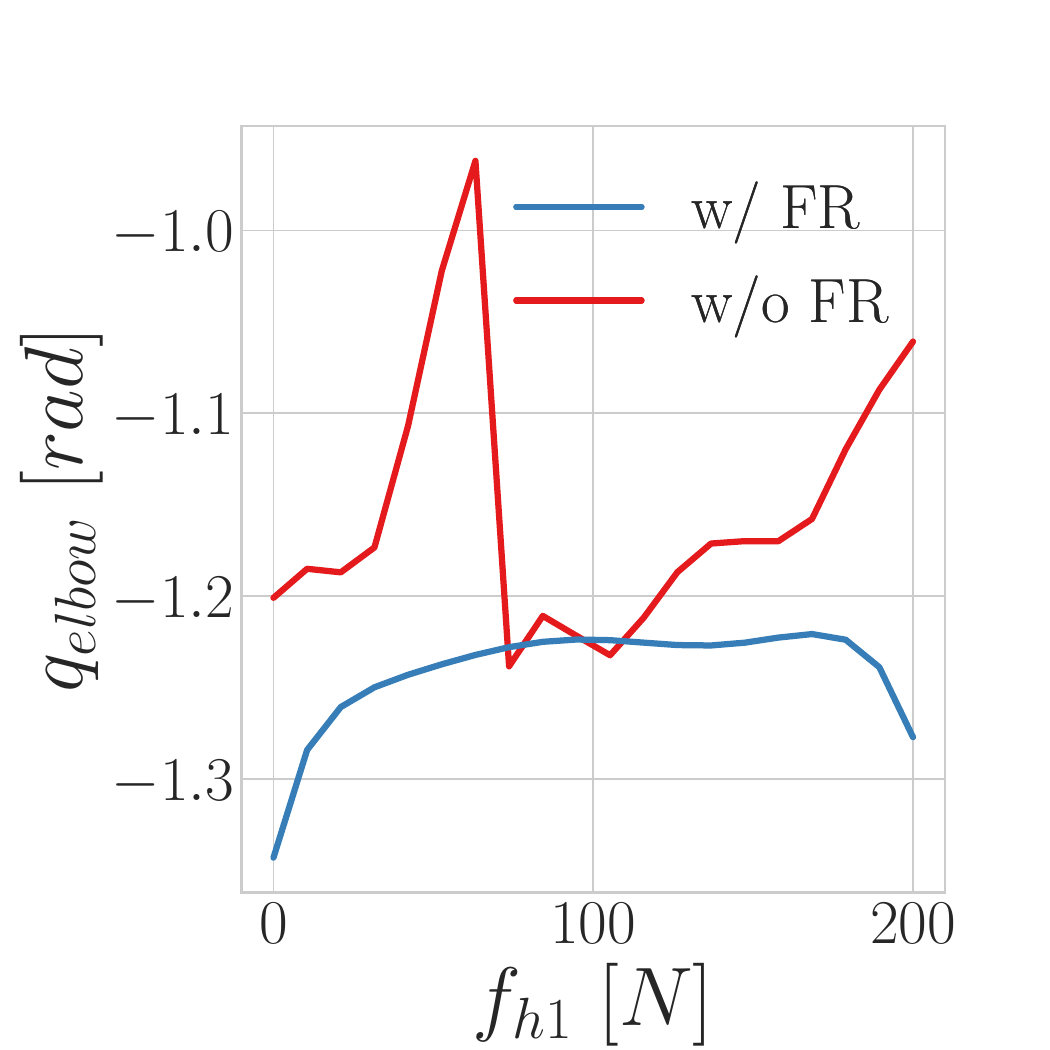}&
        \includegraphics[width = 0.45\columnwidth]{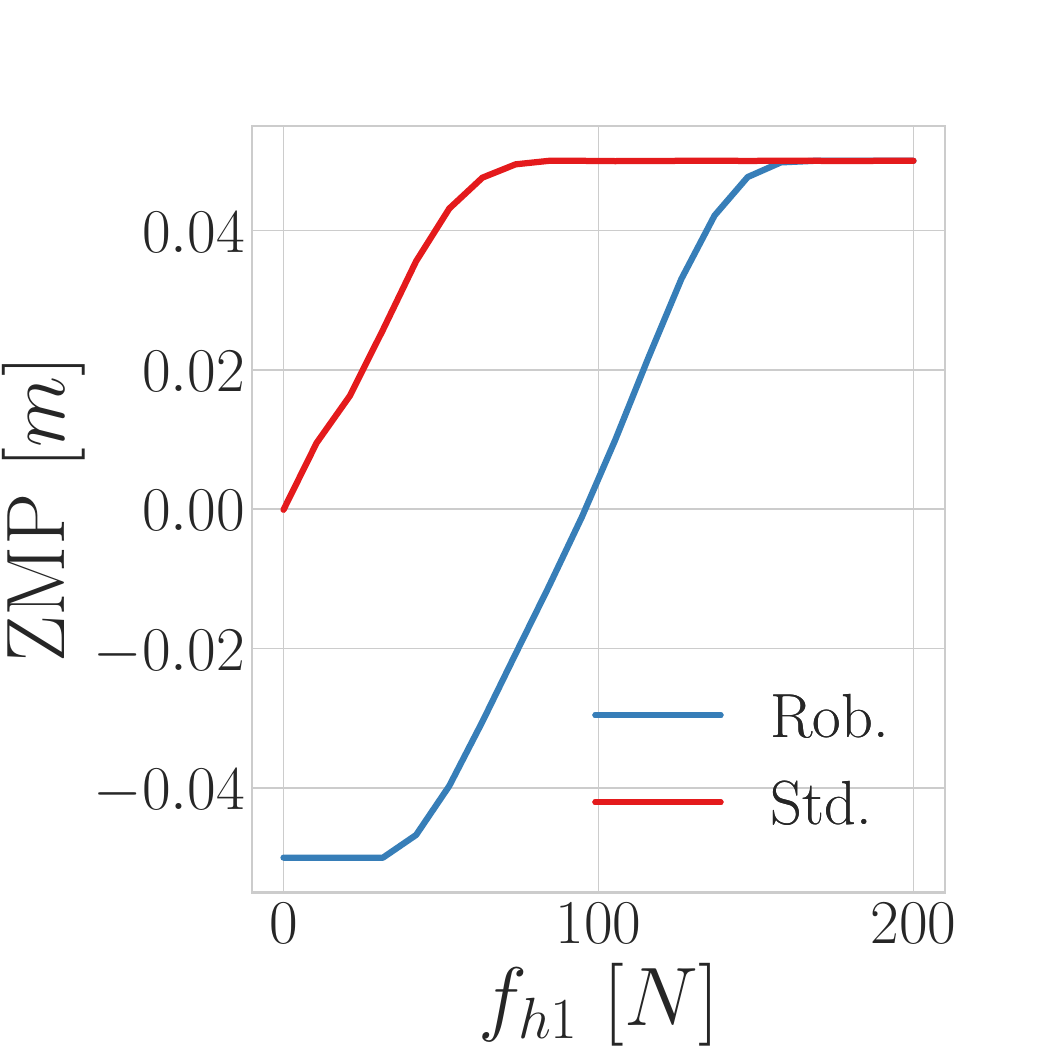} \\
        a) Elbow joint angles  & b) ZMP point 
    \end{tabular}     
    \captionof{figure}{Exploring the impact of diverse steps on varying force magnitudes.}\vspace{-5mm}
    \label{fig:different_steps}
\end{center}
\vspace{-0.4cm}
\end{table}

\begin{equation}
     \bm{q}(s) = \bm{\Psi}(s) \bm{w}, \ \text{with} \ 
     s = \ell(f_{\text{h1}}),
 \end{equation} 
where a function $\ell(\cdot)$ maps the force to the value of $s$. In this formulation, $\bm{q}$ can be regarded as a spatial reference, while the mapping variable $s$ can be regarded as a latent variable that determines the position along the reference. In this work, we assume a simple linear relationship between $s$ and the applied force, i.e., $s = \frac{f_{\text{h1}}}{f_{\text{max}}}$, and leave it for future studies to develop an enhanced controller that operates effectively on this one-dimensional manifold. Now, the problem consists of finding the desired coefficients $\bm{w}$ for the entire task, which can be solved by minimizing the optimization problem 
 \begin{equation}\label{eq:final_eq_simp}
 \begin{gathered}
    \bm{w}^* = \underset{\bm{w}}{\text{argmin}} \sum_{i=0}^{i = D} c_{\text{rob}}(\bm{q}(s_i)), \\
    \text{s.t.} \ \ \bm{q}(s_i) \in \mathcal{Q}^{\text{active}}, \ \ \text{ZMP}(\bm{q}(s_i), f_{\text{h1}}(s_i)) \in \delta^{\pm}, \\ 
    f_{\text{h1}}(s_i) = sf_{\text{max}}, \ \ \ g(\bm{q}(s), f_{\text{h1}}(s)) \leq 0,
 \end{gathered}
 \end{equation}
where $s_i = \frac{i}{D}$ for $i \in \{0,1,2,\hdots,D\}$. The summation in \eqref{eq:final_eq_simp} is going to be solved for $D+1$ discrete points on the manifold.

\section{Experiment}
\label{sec:exp}
This section analyses the efficacy of the proposed method through both simulation and real-world experimentation by conducting two experiments. The first experiment, which we refer to as the \emph{robust case}, has utilized $c_{\text{rob}}$ to find a robust solution for a range of force values, while the second experiment considers only the current force values, i.e., $c = c_{\text{std}}$, and is referred to as the \emph{standard case}. The videos presenting all the experiments are accessible through the supplementary materials or via \url{https://www.youtube.com/watch?v=GQAad6GFPlE}

\subsection{Finding the Spatial Reference}
\begin{figure}[tb]
     \centering
     \includegraphics[width = 0.8 \columnwidth]{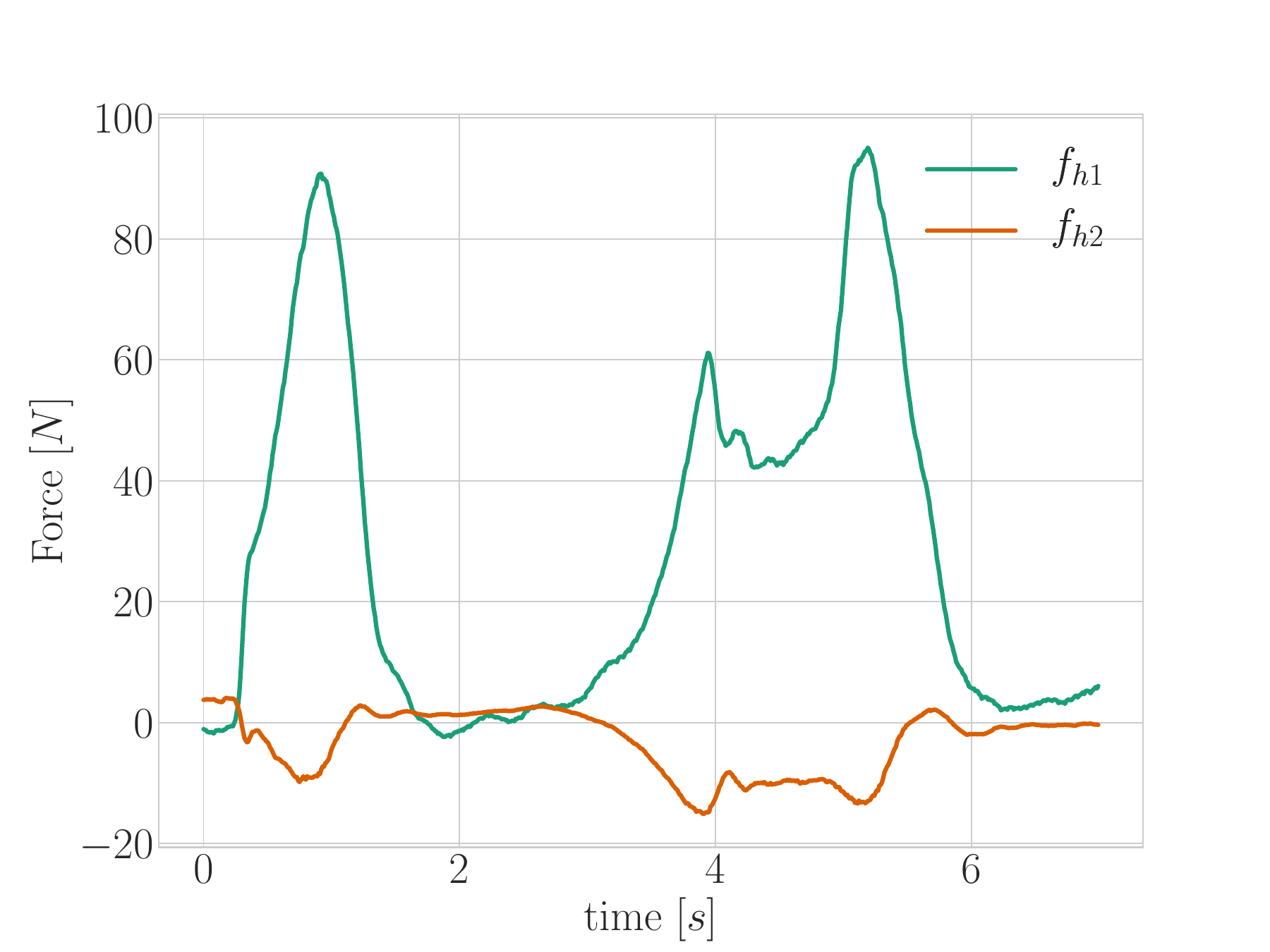}
     \caption{The force that the user applied at one of his/her hands when interacting with another user.}\vspace{-5mm}
     \label{fig:force_hh}
 \end{figure}

The initial step in this study is to solve \eqref{eq:final_eq_simp} for the robotic system. Based on the data gathered from human-human interaction (Fig. \ref{fig:force_hh}), it has been determined that a force of approximately $90\, \text{N}$ is required at each hand to enable standing up. Thus, the robot must be capable of withstanding forces up to $180\, \text{N}$ to replicate the motion. In consideration of safety concerns, a maximum value of $f_{\text{max}} = 200\,\text{N}$ has been selected. Although the ZMP for Talos can fall within a range of $\pm 10\,\text{cm}$ around the foot, a conservative margin of $\delta=5\,\text{cm}$ has been incorporated to account for unmodeled factors such as friction and to accommodate the simplifying assumptions utilized in this study. Additionally, to ensure safety, the system has been constrained to not exceed a hand position displacement of $10\,\text{cm}$ from its default configuration.

B\'ezier curves with 11 control points are utilized in this study (number selected empirically). Moreover, to exploit the symmetrical characteristics of the 14 active joints, the coordination matrix $\bm{C}$ has been defined as
\begin{equation}
    \bm{C}_{14 \times 7} = \begin{bmatrix} 1 & 1 & 0 & 0 & \hdots & 0 & 0 \\
                            0 &  0 & 1 & 1 & \hdots & 0 & 0 \\
                            \vdots & \vdots & \vdots & \vdots & \ddots & \vdots& \vdots \\
                            0 & 0 & 0 & 0 & \hdots & 1 & 1 
                \end{bmatrix}^\trsp.
\end{equation}

We have solved \eqref{eq:final_eq_simp} offline by discretizing $s$ into 20 equally spaced points. The resulting elbow joint and ZMP values are illustrated in Fig.~\ref{fig:different_steps}. We have used the SLSQP optimization method, available in the SciPy package \cite{scipy} for solving the optimization problem.
\subsection{Results}
\subsubsection{Simulation} The simulation was executed using a ROS interface on the Docker file provided by PAL Robotics. Gazebo was utilized for physical simulation, with a PD position controller applied to the robot's motor current. It should be noted that the method described here is not restricted to a position controller and that alternative controllers could also be utilized. The control system reads the force values applied to the robot, calculates the desired latent position on the manifold and its corresponding active joint configurations, and publishes the desired values at 20 Hz.

To replicate the human behaviors in the simulation, a sinusoidal force was applied for $2h$ seconds, reaching its maximum value of $M \in [100, 200]\,\text{N}$ within $h \in [1, 4]$ seconds. The results of this simulation are presented in Fig.~\ref{fig:sim_random_cases}. The findings suggest that the robust version of the system demonstrates a significantly higher success rate in handling the challenges inherent to the STS task. It is worth noting that the spatial reference generated for this experiment assumed static motion, which may not hold true during the experiment. This underlying assumption may have contributed to the system failure in certain instances. However, the system developed by taking into account the unpredictability of human behavior demonstrates greater resilience and produces more robust spatial references.

\begin{table}[t]
\vspace{2mm}
\begin{center}
    \begin{tabular}{cc}
        \centering
        \includegraphics[width =0.45\columnwidth]{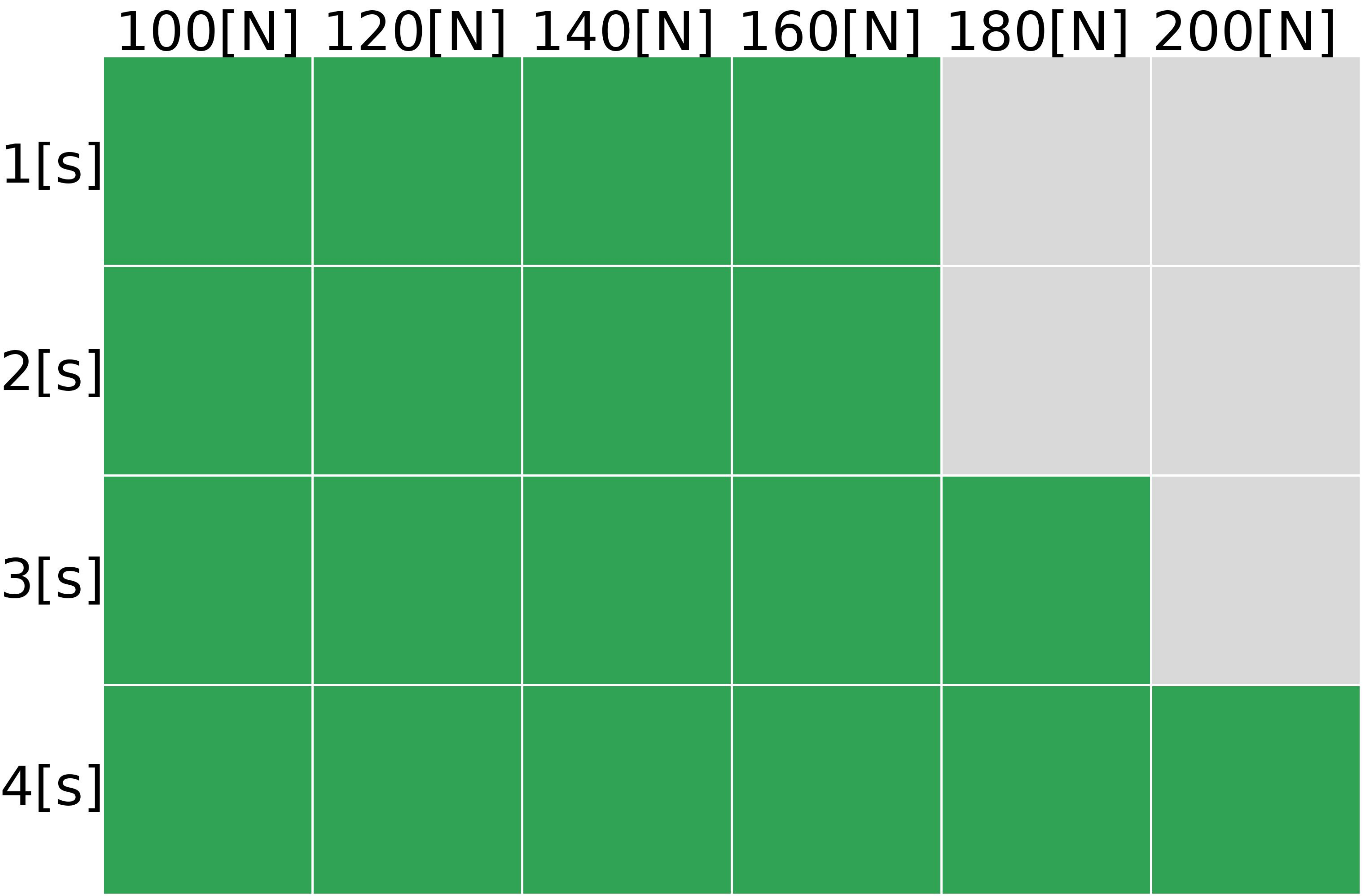}&
      \includegraphics[width =0.45\columnwidth]{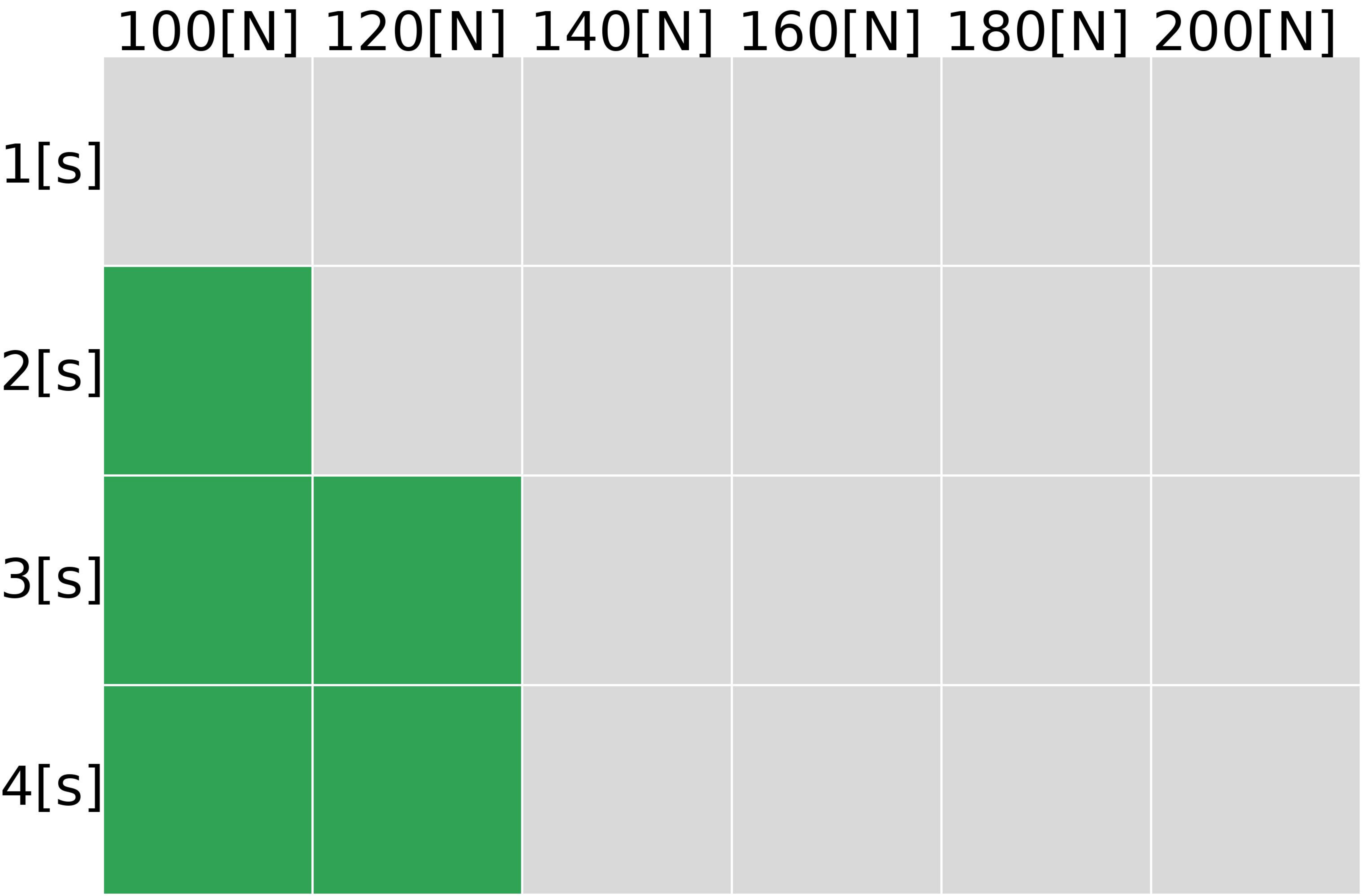}\\
        a) Robust & b) Standard\\
    \end{tabular}     
    \captionof{figure}{Examining the performance of the two systems operating under different force magnitudes $M \ [N]$ (columns), and its rising time $h \ [s]$ (rows). The force is applied for $2h$ seconds. Green and gray blocks represent successful and failure motions, respectively.}\vspace{-3mm}
    \label{fig:sim_random_cases}
\end{center}
\vspace{-0.5cm}
\end{table} 

\subsubsection{Real Robot} Following the verification of results in the simulation, a real-world experiment was conducted to validate the system performance in a practical setting. Participants were asked to perform the STS task on the robot, and the same setup was employed as in the simulation, with the only difference being the replacement of the Gazebo simulator with the real robot.

The force values applied by the participants on the robot are depicted in Fig. \ref{fig:force_real}, indicating that the maximum assumed force was not exceeded. Figure \ref{fig:force_real} also illustrates how the ZMP point on the robot varied during the real-world experiment. Although the controller was derived under the static assumption, the system attempted to remain within the specified boundary. Nevertheless, in some instances, the ZMP deviated from the marginal boundary $\delta$ while remaining within the stable region. Although the control mechanism itself was not capable of precisely controlling the ZMP point, the manifold derived by the robust case displayed appropriate resilience in preventing the system from going beyond the marginal boundary and stable region.

\begin{table*}[t]
\vspace{3mm}
\begin{center}
    \begin{tabular}{cccc}
        \centering
        \includegraphics[width=0.5\columnwidth]{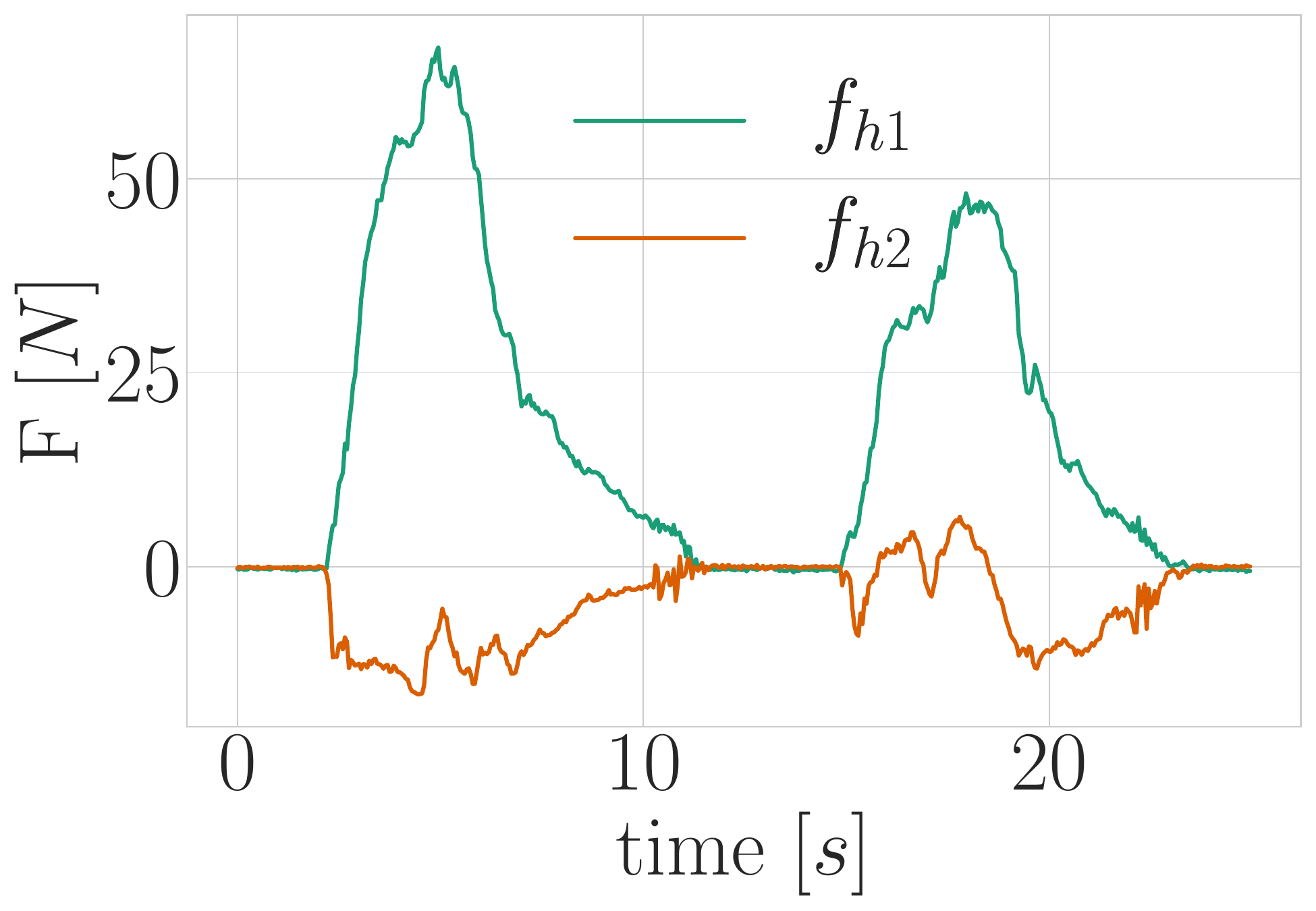}&
        \includegraphics[width=0.45\columnwidth]{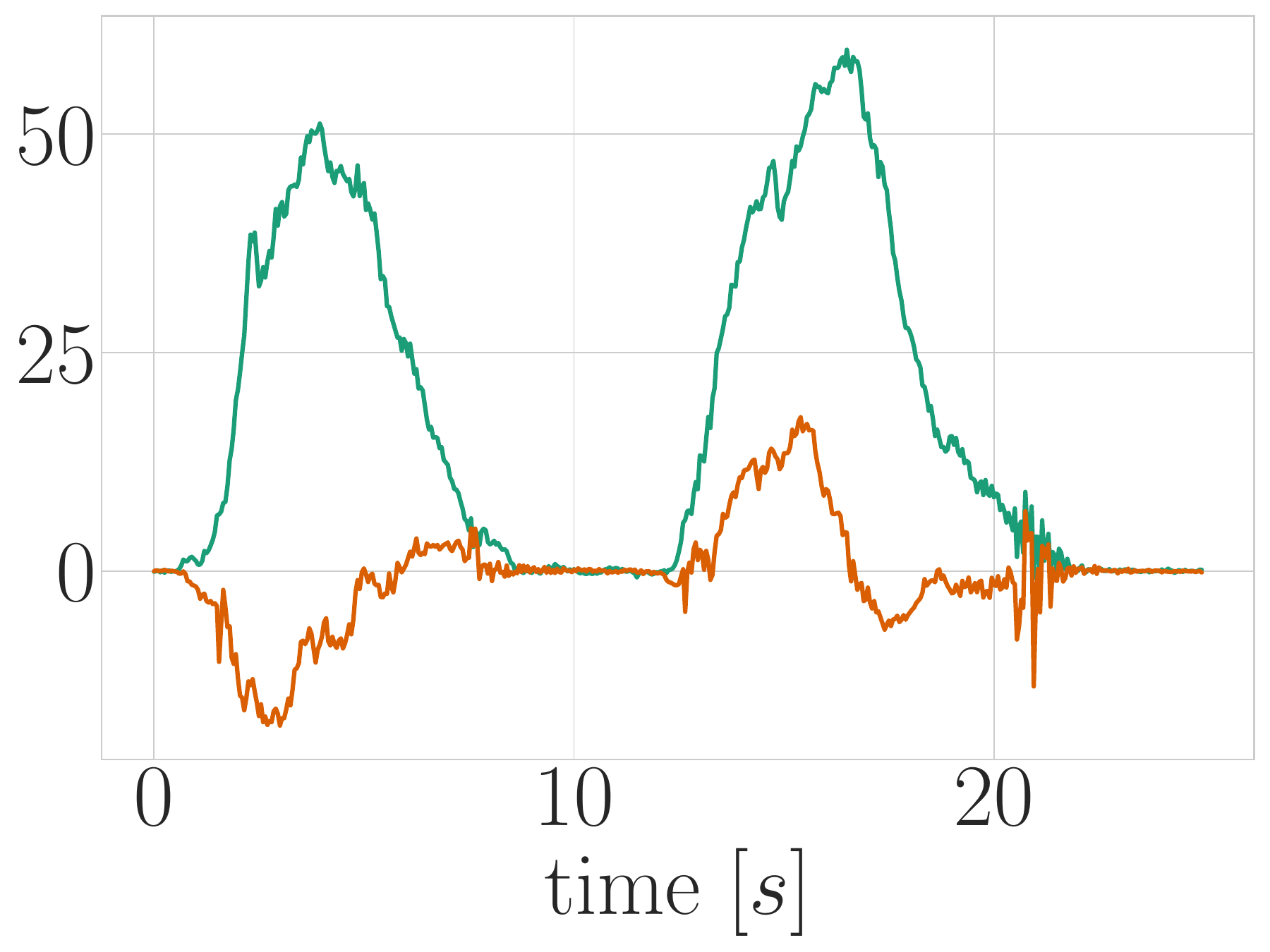}& 
        \includegraphics[width=0.45\columnwidth]{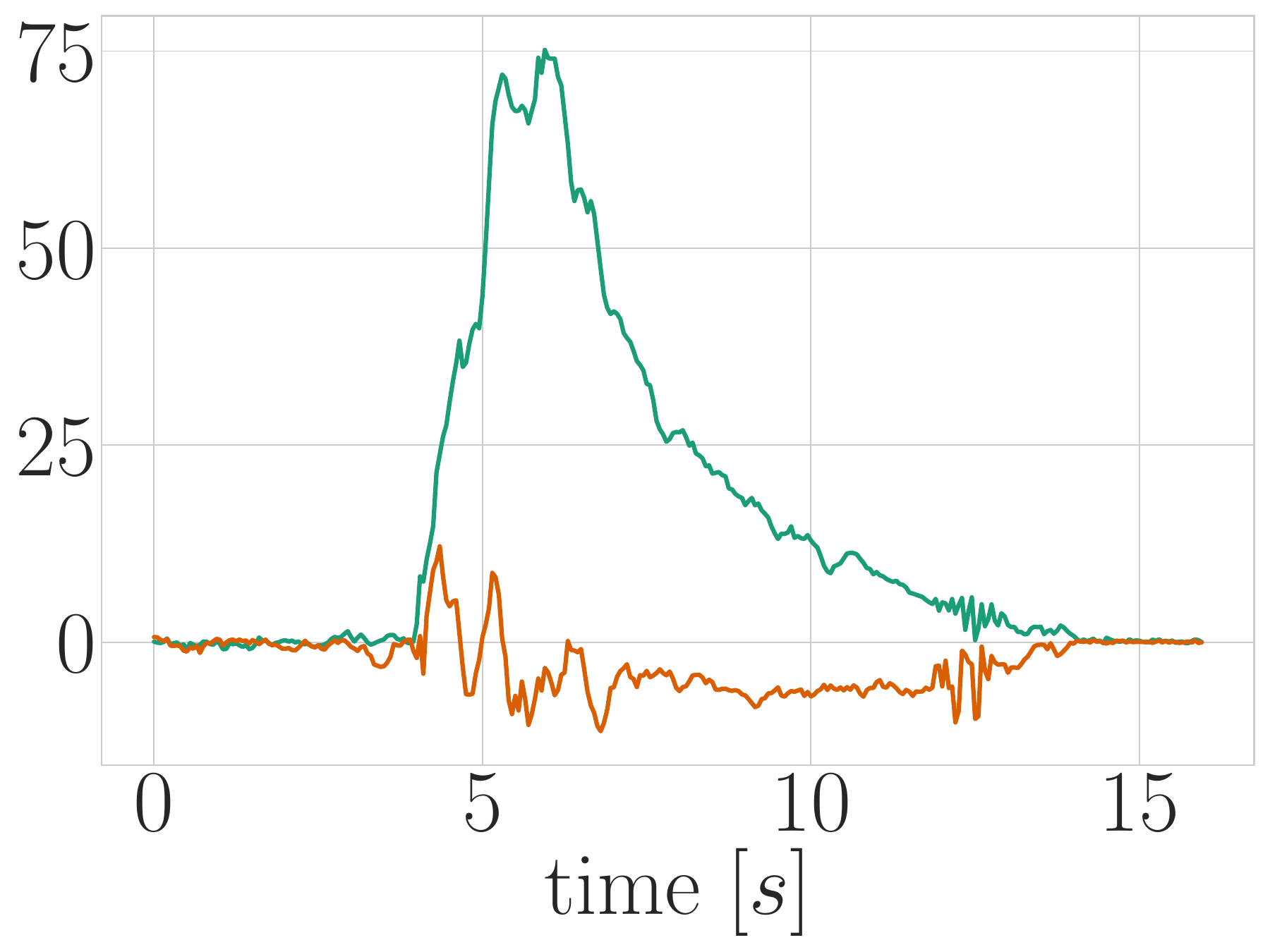}& 
        \includegraphics[width=0.45\columnwidth]{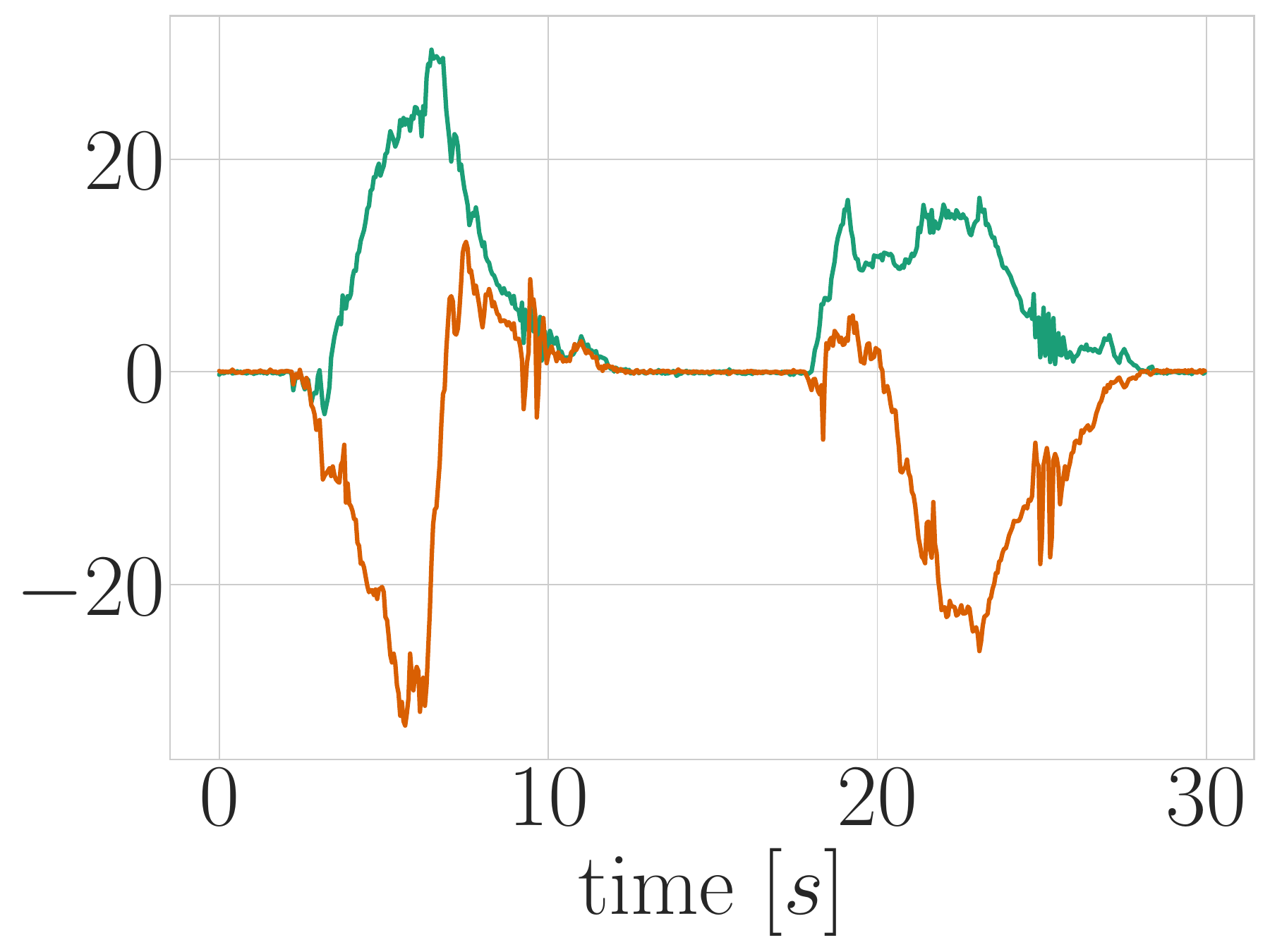} \\
        \includegraphics[width=0.5\columnwidth]{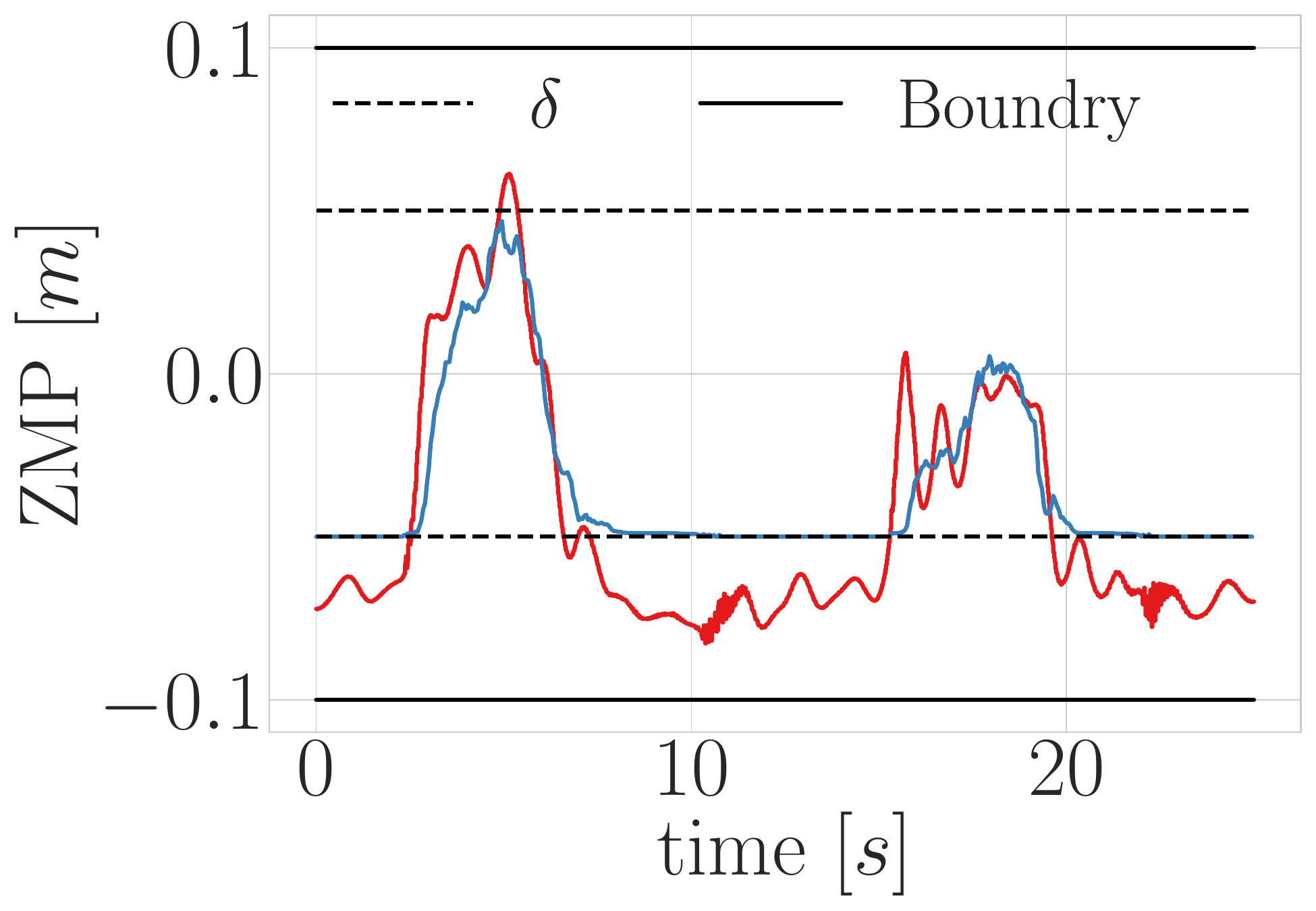}&  
        \includegraphics[width=0.45\columnwidth]{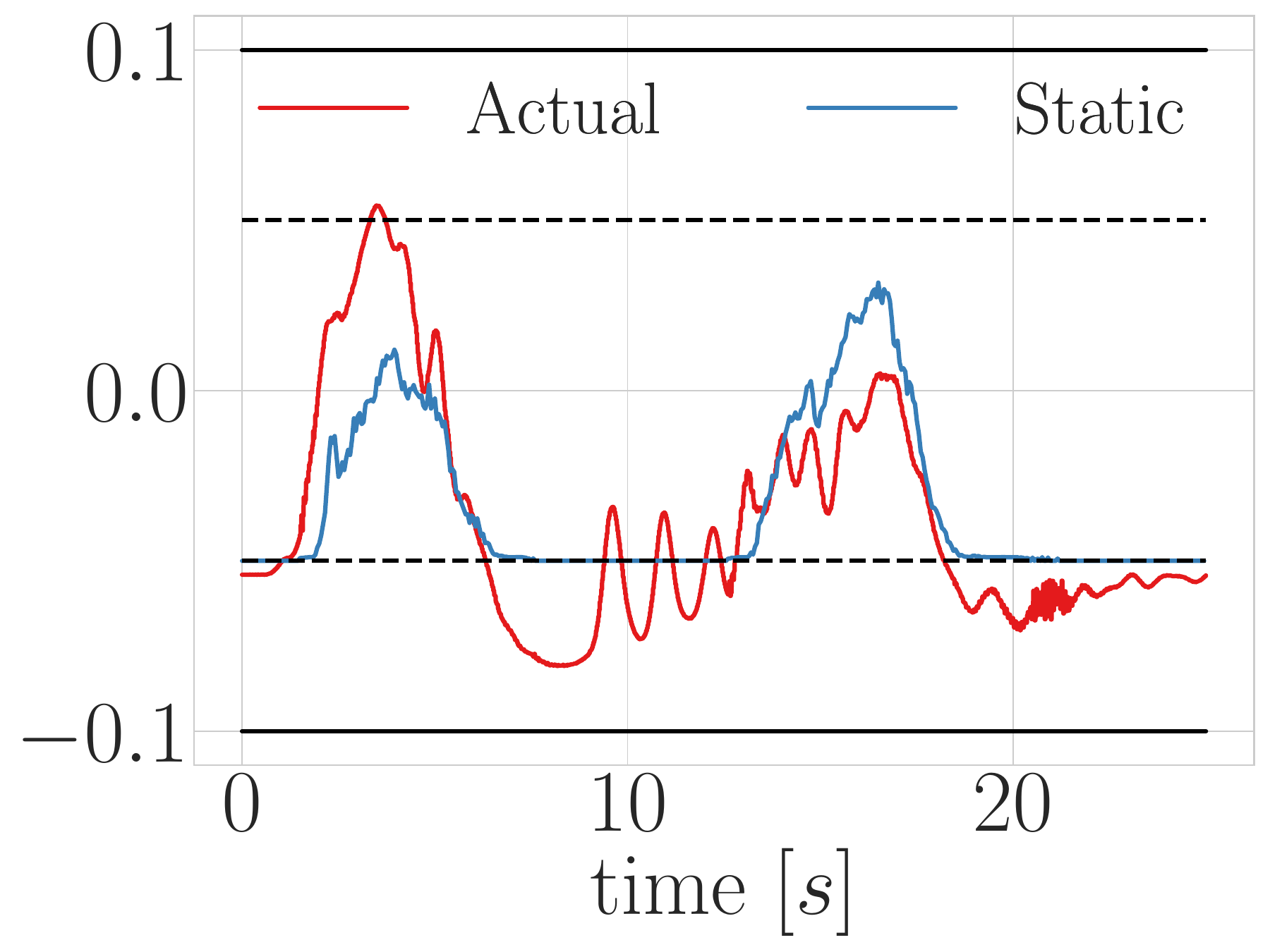}& 
        \includegraphics[width=0.45\columnwidth]{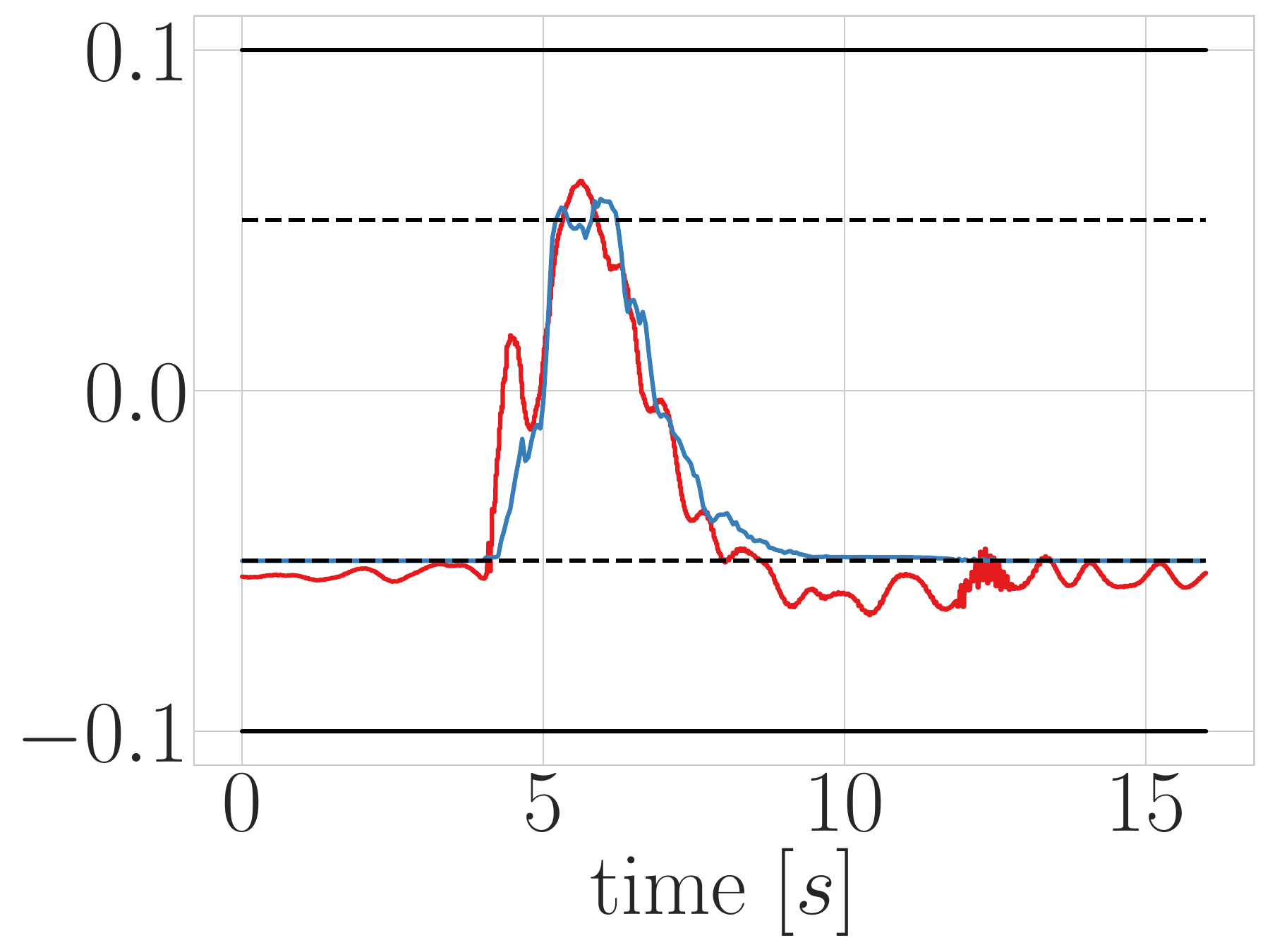}& 
        \includegraphics[width=0.45\columnwidth]{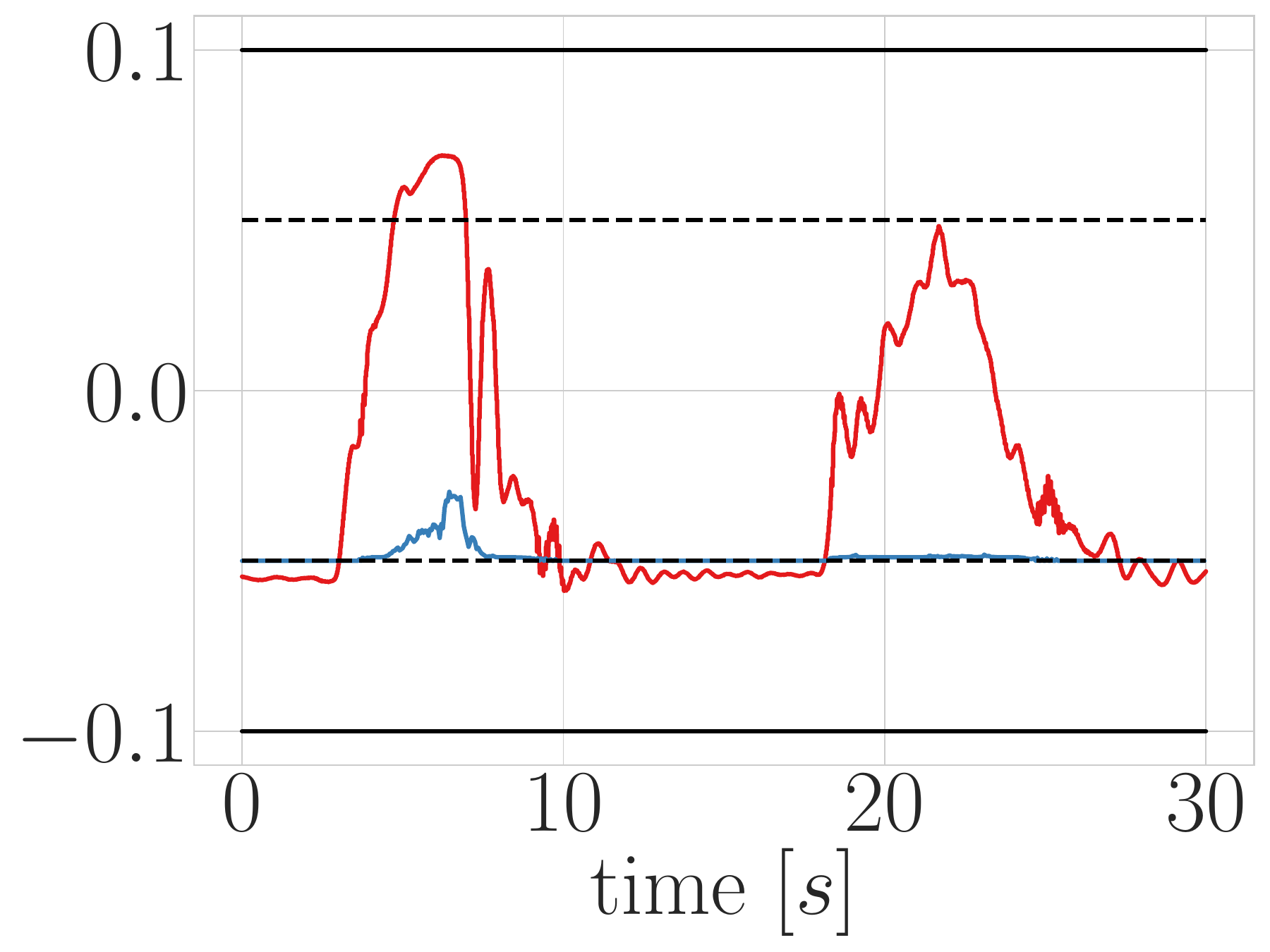}\\
    \end{tabular}     
    \captionof{figure}{The forces users applied at each robot's hand (the first row), and their corresponding ZMP change (the second row). Different columns correspond to different users.}  
    \label{fig:force_real} \vspace{-6mm}
\end{center}
\end{table*}

\section{Discussion}
\label{sec:discussion}
This section discusses the effects of the steps taken to solve the STS task using the proposed method. 
Without considering the use of a functional representation, there could be big discontinuities in the joint space, as shown in Fig. \ref{fig:different_steps}.a with jumps up to $0.2$ rad ($\approx 11.46$ degrees) for the elbow joint. We observed that a functional representation eliminates the jump in the spatial reference by correlating the optimization solution for all force values using the location-independent variables $\bm{w}$.  

We incorporated human unpredictability into the cost function by optimizing it over the range of the expected human behavior. The results, as illustrated in Fig. \ref{fig:different_steps}.b, demonstrate that our approach allows the robot to utilize the marginal boundary more effectively. Specifically, at the beginning of the task, the robot anticipates the possibility of a force being applied by the human and adjusts its ZMP point accordingly to provide a larger safety margin. Conversely, the standard method focuses solely on regulating the ZMP point of each force around zero. The advantages of this robustness are demonstrated in Fig.~\ref{fig:sim_random_cases}. To further enhance this approach, future studies may consider weighting each potential human behavior by its occurrence probability. This would enable the system to better adapt to more likely scenarios, and potentially lead to more personalized and efficient task performance.

The method that we propose has some remaining limitations. Fig. \ref{fig:force_real} shows that the ZMP points slightly violate the marginal boundaries $\delta$. This is because the method does not have direct control over the ZMP point and relies on the robustness of the manifold. This error can also be related to unmodeled dynamics such as the friction effect, especially when there is no external force, where ZMP stays around $-0.07\,\text{cm}$ instead of $-0.05\,\text{cm}$. In future work, we will investigate adding feedback on the ZMP point to control it more directly on the manifold. Another idea could be to solve \eqref{eq:main_problem} on the manifold, removing the need to consider some constraints such as joint limits, and optimizing a one-dimensional variable $\ddot{s}$ over a horizon. The efficiency of these methods should be investigated in future studies.

One observation from the experimental results is that for the user shown in the rightmost column in Fig. \ref{fig:force_real}, the estimated ZMP point using a static assumption differs significantly from the actual one, despite the successful completion of the task. This phenomenon is also evident in other experiments (not depicted here), particularly when the user applies forces below 40 N on each hand (80 N on the robot). Our preliminary hypothesis is that, as shown in Fig. \ref{fig:different_steps}.b when the force is below 40 N (20 N at each hand), the robot's movement is concentrated on maintaining the ZMP on the boundary, which requires agile motion of CoM to compensate the external force. Consequently, the static assumption is more likely to be violated in this phase. This can also be seen by comparing the amount of motion in Fig. \ref{fig:different_steps}.a at the beginning of the task versus the remainder of the task. Additionally, since the robot's motion is controlled through a position controller, the accuracy of the calculated ZMP depends on the controller's precision in tracking the reference motion, which is expected to be less accurate for agile motions. To address this issue, we can either modify the underlying controller or incorporate an online controller. Furthermore, unlike the others, this user has applied force of comparable or greater magnitude in the vertical direction, as opposed to the horizontal direction. We initially assumed that the impact of this force would be negligible compared to the effect of the horizontal force. However, for this particular user, this assumption may not hold true. This outcome may be attributed to the non-adaptive initial pose of the robot's hand, which might not have been comfortable for certain users, resulting in non-intuitive force patterns. While this force was insufficient to cause the robot to fall, we are motivated to consider both forces in our future formulation, which will result in higher dimensional manifolds.
\section{CONCLUSIONS}
\label{sec:conclude}
We demonstrated how the concept of the manifold can be leveraged to effectively utilize the redundancy of a problem to construct an interpretable low-dimensional latent space for the robot. This manifold can be interpreted as local coordination among the system, which correlates all of the joints in accordance with the desired position of the robot on the manifold. We analyzed the impact of different stages of the proposed method on the final results. While incorporating the functional representation idea leads to a smooth manifold, considering the uncertainty associated with human behavior enables the system to exploit the redundancy in a broader way.

In this study, we assumed a simple controller that maps the force value to the position of the robot on the manifold and does not address online controller design. Future work could involve the use of more sophisticated controllers that would update the desired location based on various parameters such as the zero moment point. It may be possible to utilize a fast model predictive controller (MPC) control strategy acting on this 1D manifold. Furthermore, an area for improvement is optimization under uncertain human behaviors. This can be achieved through various methods of predicting human behavior and weighting different behaviors in the cost function according to their likelihood of occurrence. Moreover, extending the concept of manifold to multidimensional manifolds may lead to interesting results. Lastly, an intriguing avenue to pursue would involve jointly optimizing over both the predefined coordination matrix $\bm{C}$ and the superposition weights $\bm{w}$. Such an approach could potentially generate even more significant synergies between the joints, thereby offering additional opportunities for the controller to exploit.






\bibliographystyle{IEEEtran}
\bibliography{root}

\begin{thebibliography}{10}
\providecommand{\url}[1]{#1}
\csname url@samestyle\endcsname
\providecommand{\newblock}{\relax}
\providecommand{\bibinfo}[2]{#2}
\providecommand{\BIBentrySTDinterwordspacing}{\spaceskip=0pt\relax}
\providecommand{\BIBentryALTinterwordstretchfactor}{4}
\providecommand{\BIBentryALTinterwordspacing}{\spaceskip=\fontdimen2\font plus
\BIBentryALTinterwordstretchfactor\fontdimen3\font minus
  \fontdimen4\font\relax}
\providecommand{\BIBforeignlanguage}[2]{{%
\expandafter\ifx\csname l@#1\endcsname\relax
\typeout{** WARNING: IEEEtran.bst: No hyphenation pattern has been}%
\typeout{** loaded for the language `#1'. Using the pattern for}%
\typeout{** the default language instead.}%
\else
\language=\csname l@#1\endcsname
\fi
#2}}
\providecommand{\BIBdecl}{\relax}
\BIBdecl

\bibitem{rasouli_potential_2022}
S.~Rasouli, G.~Gupta, E.~Nilsen, and K.~Dautenhahn, ``Potential applications of
  social robots in robot-assisted interventions for social anxiety,'' vol.~14,
  no.~5, pp. 1--32.

\bibitem{Scassellati_2022}
M.~Qin, J.~Brawer, and B.~Scassellati, ``Task-oriented robot-to-human handovers
  in collaborative tool-use tasks,'' in \emph{31st IEEE International
  Conference on Robot and Human Interactive Communication (RO-MAN)}, 2022, pp.
  1327--1333.

\bibitem{Ajoudani_2022}
D.~Sirintuna, A.~Giammarino, and A.~Ajoudani, ``Human-robot collaborative
  carrying of objects with unknown deformation characteristics,'' in
  \emph{IEEE/RSJ International Conference on Intelligent Robots and Systems
  (IROS)}, 2022, pp. 10\,681--10\,687.

\bibitem{Geravand2017}
M.~Geravand, P.~Z. Korondi, C.~Werner, K.~Hauer, and A.~Peer, ``{Human
  sit-to-stand transfer modeling towards intuitive and biologically-inspired
  robot assistance},'' \emph{Autonomous Robots}, vol.~41, no.~3, pp. 575--592,
  2017.

\bibitem{Razmjoo21ICAR}
A.~Razmjoo, T.~S. Lembono, and S.~Calinon, ``Optimal control combining
  emulation and imitation to acquire physical assistance skills,'' in
  \emph{Proc.\ {IEEE} Intl Conf.\ on Advanced Robotics ({ICAR})}, 2021, pp.
  338--343.

\bibitem{Li2021}
J.~Li, L.~Lu, L.~Zhao, C.~Wang, and J.~Li, ``{An integrated approach for
  robotic Sit-To-Stand assistance: Control framework design and human intention
  recognition},'' \emph{Control Engineering Practice}, vol. 107, p. 104680,
  2021.

\bibitem{shimon_STS_2015}
M.~Shomin, J.~Forlizzi, and R.~Hollis, ``Sit-to-stand assistance with a
  balancing mobile robot,'' in \emph{2015 IEEE International Conference on
  Robotics and Automation (ICRA)}, pp. 3795--3800.

\bibitem{Kajita1991}
S.~Kajita and K.~Tani, ``Study of dynamic biped locomotion on rugged
  terrain-derivation and application of the linear inverted pendulum mode,'' in
  \emph{Proc. {IEEE} Intl Conf. on Robotics and Automation ({ICRA})}, pp.
  1405--1411.

\bibitem{dai2014whole}
H.~Dai, A.~Valenzuela, and R.~Tedrake, ``Whole-body motion planning with
  centroidal dynamics and full kinematics,'' in \emph{Proc. {IEEE} Intl Conf.
  on Humanoid Robots ({H}umanoids)}, 2014, pp. 295--302.

\bibitem{Atkeson2007}
C.~G. Atkeson and B.~Stephens, ``Multiple balance strategies from one
  optimization criterion,'' in \emph{Proc. {IEEE} Intl Conf. on Humanoid Robots
  ({H}umanoids)}, 2007, pp. 57--64.

\bibitem{Ibanez2018}
A.~Ibanez, P.~Bidaud, and V.~Padois, ``Optimization-based control approaches to
  humanoid balancing,'' in \emph{Humanoid Robotics: A Reference}.\hskip 1em
  plus 0.5em minus 0.4em\relax Springer Netherlands, pp. 1--27.

\bibitem{Ibanez2012}
------, ``Unified preview control for humanoid postural stability and
  upper-limb interaction adaptation,'' in \emph{Proc. {IEEE/RSJ} Intl Conf. on
  Intelligent Robots and Systems ({IROS})}, pp. 1801--1808.

\bibitem{orin_centroidal_2013}
D.~E. Orin, A.~Goswami, and S.-H. Lee, ``Centroidal dynamics of a humanoid
  robot,'' \emph{Autonomous robots}, vol.~35, pp. 161--176, 2013.

\bibitem{Hirukawa_2006}
H.~Hirukawa, S.~Hattori, K.~Harada, S.~Kajita, K.~Kaneko, F.~Kanehiro,
  K.~Fujiwara, and M.~Morisawa, ``A universal stability criterion of the foot
  contact of legged robots - adios zmp,'' in \emph{Proceedings IEEE
  International Conference on Robotics and Automation (ICRA)}, 2006, pp.
  1976--1983.

\bibitem{Ijspeert2013}
A.~J. Ijspeert, J.~Nakanishi, H.~Hoffmann, P.~Pastor, and S.~Schaal,
  ``{Dynamical movement primitives: Learning attractor models formotor
  behaviors},'' \emph{Neural Computation}, vol.~25, no.~2, pp. 328--373, 2013.

\bibitem{paraschos_probabilistic_2013}
A.~Paraschos, C.~Daniel, J.~R. Peters, and G.~Neumann, ``Probabilistic movement
  primitives,'' in \emph{Advances in Neural Information Processing Systems},
  C.~J. Burges, L.~Bottou, M.~Welling, Z.~Ghahramani, and K.~Q. Weinberger,
  Eds., vol.~26.\hskip 1em plus 0.5em minus 0.4em\relax Curran Associates,
  Inc., 2013.

\bibitem{Calinon19MM}
S.~Calinon, ``Mixture models for the analysis, edition, and synthesis of
  continuous time series,'' in \emph{Mixture Models and Applications},
  N.~Bouguila and W.~Fan, Eds.\hskip 1em plus 0.5em minus 0.4em\relax Springer,
  Cham, 2019, pp. 39--57.

\bibitem{FAROUKI2012379}
R.~T. Farouki, ``The bernstein polynomial basis: A centennial retrospective,''
  \emph{Computer Aided Geometric Design}, vol.~29, no.~6, pp. 379--419, 2012.

\bibitem{wieber2016modeling}
P.-B. Wieber, R.~Tedrake, and S.~Kuindersma, ``Modeling and control of legged
  robots,'' in \emph{Springer handbook of robotics}.\hskip 1em plus 0.5em minus
  0.4em\relax Springer, 2016, pp. 1203--1234.

\bibitem{scipy}
``Scipy 1.0: fundamental algorithms for scientific computing in python,''
  \emph{Nature Methods}, vol.~17, pp. 261--272, 2020.

\end{thebibliography}

\end{document}